\documentclass[11pt, a4paper, logo, twocolumn, copyright]{deepmind} 
\usepackage[mathletters]{ucs} 
\usepackage[utf8x]{inputenc} 
\usepackage[linesnumbered, ruled]{algorithm2e}
\SetAlCapSkip{1em}
\SetKwInput{KwParam}{Parameters}
\SetKwInput{KwHyper}{Hyperparameters}

\SetCommentSty{mycommfont}
\usepackage{cleveref}

\def\citep{\cite}
\def\citet{\cite}
\def\v#1{{\boldsymbol{#1}}} 
\def\m#1{{\boldsymbol{#1}}} 
\def\t{\mathrm} 
\def\vth{{\boldsymbol\theta}} 
\def\vbe{{\boldsymbol\beta}} 
\def\vga{{\boldsymbol\gamma}} 
\def\bmcW{{\boldsymbol{\mathcal{W}}}} 
\def\bmcWqkv{\boldsymbol{\mathcal{W}}_{\!\!\boldsymbol{qkv}}} 

\title{Formal Algorithms for Transformers}

\correspondingauthor{$\{$buiphuong,mhutter$\}$@deepmind.com}
\renewcommand{\today}{19 July 2022}
\author[1]{Mary Phuong}
\author[1]{Marcus Hutter}
\affil[1]{DeepMind}

\begin{abstract}
This document aims to be a self-contained, mathematically precise overview of transformer architectures and algorithms (\emph{not} results).
It covers what transformers are, how they are trained, what they are used for, their key architectural components, and a preview of the most prominent models.
The reader is assumed to be familiar with basic ML terminology and simpler neural network architectures such as MLPs.
\end{abstract}

\keywords{formal algorithms, pseudocode, transformers, attention, encoder, decoder, BERT, GPT, Gopher, tokenization, training, inference.}

\begin{document}
\maketitle

\def\contentsname{\centering\normalsize Contents}\setcounter{tocdepth}{1}
{\parskip=-2.5ex\tableofcontents}\vspace{3ex}

{\it A famous colleague once sent an actually very well-written paper he was quite proud of to a famous complexity theorist. His answer: ``I can't find a theorem in the paper. I have no idea what this paper is about.''}

\section{Introduction}\label{sec:Intro}

Transformers are deep feed-forward artificial neural networks with a (self)attention mechanism.
They have been tremendously successful in natural language processing tasks and other domains.
Since their inception 5 years ago \citep{vaswani17}, many variants have been suggested \citep{Lin:21}.
Descriptions are usually graphical, verbal, partial, or incremental.
Despite their popularity, it seems no pseudocode has ever been published for any variant.
Contrast this to other fields of computer science, 
even to ``cousin'' discipline reinforcement learning \cite{Mnih:13,Sutton:18,Efroni:21}.

This report intends to rectify the situation for Transformers.
It aims to be a self-contained, complete, precise and compact overview of transformer architectures and formal algorithms (but \emph{not} results).
It covers what Transformers are (\Cref{sec:architectures}), 
how they are trained (\Cref{sec:train}), 
what they're used for (\Cref{sec:tasks}), 
their key architectural components (\Cref{sec:components}), 
tokenization (\Cref{sec:token}), 
and a preview of practical considerations (\Cref{sec:practice}) and the most prominent models.

The essentially complete pseudocode is about 50 lines, compared to thousands of lines of actual real source code.
We believe these formal algorithms will 
be useful for theoreticians who require compact, complete, and precise formulations,
experimental researchers interested in implementing a Transformer from scratch,
and encourage authors to augment their paper or text book with formal Transformer algorithms (\Cref{sec:Motivation}).  

The reader is assumed to be familiar with basic ML terminology and simpler neural network architectures such as MLPs.

In short, a (formally inclined) reader, upon understanding the contents of this document, will have a solid grasp of transformers: they will be ready to read and contribute to the literature on the topic as well as implement their own Transformer using the pseudocode as templates.

\section{Motivation}\label{sec:Motivation} 

The true story above the introduction describes quite well the feeling we have when browsing many Deep Learning (DL) papers; 
unable to figure out the algorithmic suggestions exactly. 
For practitioners, the papers may be sufficiently detailed, 
but the precision needed by theoreticians is usually higher.
For reasons not entirely clear to us, the DL community seems shy of providing pseudocode for their neural network models.
Below we describe the SOTA in DL paper writing and argue for the value of formal algorithms.
The reader already convinced about their merit can without loss skip this section.

\paragraph{The lack of scientific precision and detail in DL publications.}
Deep Learning has been tremendously successful in the last 5 to 10 years with thousands of papers published every year.
Many describe only informally how they change a previous model,
Some 100+ page papers contain only a few lines of prose informally describing the model \cite{rae21}.
At best there are some high-level diagrams.
No pseudocode. No equations. No reference to a precise explanation of the model.
One may argue that most DL models are minor variations of a few core architectures,
such as the Transformer \citep{vaswani17}, 
so a reference augmented by a description of the changes should suffice.
This would be true if (a) the changes were described precisely, 
(b) the reference architecture has been described precisely elsewhere,
and (c) a reference is given to this description.
Some if not all three are lacking in most DL papers.
To the best of our knowledge no-one has even provided pseudocode 
for the famous Transformer and its encoder/decoder-only variations.

\paragraph{Interfacing algorithms.}
Equally important are proper explanations of how these networks are trained and used,
but sometimes it is not even clear what the inputs and outputs and potential side-effects of the described model are.
Of course someone experienced would usually be able to correctly guess, 
but this is not a particularly scientific approach.
The experimental section in publications often does not explain what is actually fed into the algorithms and how. If there is some explanation in the methods section,
it is often disconnected from what is described in the experimental section,
possibly due to different authors writing the different sections.
The core algorithms for the models should be accompanied by the wrapper algorithms that call them, e.g.\ (pre)training, fine-tuning, prompting, inference, deployment.
Sometimes this is simple, but sometimes the magic happens there. 
In any case, if these things are not formally stated they remain unclear. 
Again, if the setup is standard and has been formally explained elsewhere,
a simple reference will do.

\paragraph{Source code vs pseudocode.}
Providing open source code is very useful, but not a proper substitute for formal algorithms. 
There is a massive difference between a (partial) Python dump and well-crafted pseudocode.
A lot of abstraction and clean-up is necessary:
remove boiler plate code, use mostly single-letter variable names,
replace code by math expressions wherever possible, e.g.\ replace loops by sums, remove (some) optimizations, etc.
A well-crafted pseudocode is often less than a page and still essentially complete,
compared to often thousands of lines of real source code.
This is hard work no-one seems to be willing to do.
Of course a forward process of first designing algorithms and write up pseudocode on paper,
and then implementing it is fine too, but few DL practitioners seem to work that way.

\paragraph{Examples of good neural network pseudocode and mathematics and explanations.}
Multi-Layer Perceptrons (MLPs) are usually well-described in many papers, e.g.\ \cite{montufar14,bartlett17,jacot18}, though also without pseudocode.
For a rare text-book example of pseudocode for a non-trivial neural network architecture, 
see Algorithm S2 of \cite{Hutter:21dgn},
which constitutes a {\em complete}, i.e.\ {\em essentially executable}, pseudocode of just 25 lines
based on a 350-line Python Colab toy implementation,
which itself is based on a proper 1000+ line implementation.

This work aims to do the same for Transformers:
The whole decoder-only Transformer GPT \Cref{algo:DTransformer} based on attention  \Cref{algo:attention,algo:MHAttention} and normalization \Cref{algo:layer_norm} including training \Cref{algo:DTraining} and prompting and inference \Cref{algo:DInference} all-together is less than 50 lines of pseudocode, 
compared to e.g.\ the 2000-line self-contained C-implementation \cite{Bellard:21}. 

\citet{Alammar:19} is a great blog-post explaining Transformers and 
\citet{Edelman:21} describes the attention mechanism to sufficient mathematical precision to allow proving properties about it, but neither provides pseudocode.
\citet{Elhage:21} is an attempt to understand Transformers by reverse-engineering the computations they perform and interpreting them as circuits.

\paragraph{Motivation.}
But does anyone actually need pseudocode and what would they be useful for (we sometimes get asked)?
We find the absence of pseudocode in DL and this question quite perplexing,
but apparently it requires answering.
Providing such pseudocode can be useful for many purposes:
\begin{itemize}\parskip=0ex\parsep=0ex\itemsep=0ex
\item They can be used as templates and adapted to precisely describe future variations, and therewith set a new standard in DL publishing.
      We explicitly encourage the reader to copy and adapt them to their needs and cite the original as ``adapted from \cite{phuong22transalg}''.
\item Having all that matters on one page in front of you makes it easier to develop new variations compared to reading prose or scrolling through 1000s of lines of actual code. 
\item They can be used as a basis for new implementations from scratch, e.g.\ in different programming languages, without having to wade through and reverse-engineer existing idiosyncratic real source code.
\item They may establish notational convention, which eases communication and reading future variations.
\item The process of converting source code into pseudocode can exhibit implementation errors (as it e.g.\ did in \cite{Hutter:21dgn}).
\item Theoreticians need compact, complete, and precise representations for reasoning and ultimately proving properties about algorithms. They are often unwilling or unable to reverse engineer code, or guess the meaning of words or fancy diagrams.
\end{itemize}

With this motivation in mind, the following five sections formally describe all aspects of transformer architectures, training, and inference.

\section{Transformers and Typical Tasks}\label{sec:tasks}  

Transformers are neural network models that excel at natural language processing, or more generally at modelling sequential data.
Two common types of tasks they are used for are \emph{sequence modelling} and \emph{sequence-to-sequence prediction}.
%
\paragraph{Notation.}
Let $V$ denote a finite set, called a \emph{vocabulary}, often identified with $[N_\t{V}]:=\{1,...,N_\t{V}\}$. 
This could be words or letters, but typically are sub-words, called tokens.
Let $\v{x}≡x[1:\ell]≡x[1]x[2]...x[\ell]∈V^*$ be a sequence of tokens, 
e.g.\ a sentence or a paragraph or a document.
Unlike in Python, we use arrays starting from $1$, 
and $x[1:\ell]$ includes $x[\ell]$.
For a matrix $M∈ℝ^{d×d'}$, we write $M[i,:]∈ℝ^{d'}$ for the $i$th row and $M[:,j]∈ℝ^d$ for the $j$-th column. 
We use matrix $×$ column vector convention more common in mathematics,
compared to the default row vector $×$ matrix in the transformer literature,
i.e.\ our matrices are transposed.
See \Cref{sec:notation} for a complete list of notation.

\paragraph{Chunking.}
The predominant paradigm in machine learning is (still) learning from independent and identically distributed (i.i.d.) data.
Even for sequence modelling for practical reasons this tradition is upheld.
The training data may naturally be a collection of (independent) articles,
but even then, some may exceed the maximal context length $\ell_{\max}$ transformers can handle.
In this case, an article is crudely broken into shorter chunks of length $≤\ell_{\max}$.

\paragraph{Sequence modelling (DTransformer).}
Given a vocabulary $V$, let $\v{x}_n∈V^*$ for $n∈[N_\t{data}]$
be a dataset of sequences (imagined to be) sampled i.i.d.\ from some distribution $P$ over $V^*$.
The goal is to learn an estimate $\hat P$ of the distribution $P(\v{x})$.
In practice, the distribution estimate is often decomposed via the chain rule as $\hat P(\v{x}) = \hat P_{\vth}(x[1]) \cdot \hat P_{\vth}(x[2] \,|\, x[1]) \cdots \hat P_{\vth}(x[\ell] \,|\, \v{x}{[1:\ell-1]})$, where $\vth$ consists of all neural network parameters to be learned.
The goal is to learn a distribution over a single token $x[t]$ given its preceding tokens $x[1:t-1]$ as context.

Examples include e.g.\ language modelling, RL policy distillation, or music generation.

\paragraph{Sequence-to-sequence (seq2seq) prediction (EDTransformer).}
Given a vocabulary $V$ and an i.i.d.\ dataset of sequence pairs $(\v{z}_n, \v{x}_n) \sim P$, where $P$ is a distribution over $V^*×V^*$, learn an estimate of the conditional distribution $P(\v{x}| \v{z})$.
In practice, the conditional distribution estimate is often decomposed as $\hat P(\v{x}| \v{z}) = \hat P_{\vth}(x[1] \,|\, \v{z}) \cdot \hat P_{\vth}(x[2] \,|\, x[1], \v{z}) \cdots \hat P_{\vth}(x[\ell] \,|\, \v{x}{[1:\ell-1]}, \v{z})$.

Examples include translation ($\v{z} =$ a sentence in English, $ \v{x} =$ the same sentence in German), question answering ($ \v{z} =$ question, $ \v{x}=$ the corresponding answer), text-to-speech ($ \v{z} =$ a piece of text, $ \v{x} =$ a voice recording of someone reading the text).

\paragraph{Classification (ETransformer).}
Given a vocabulary $V$ and a set of classes $[N_\t{C}]$, let $(\v{x}_n, c_n) \in V^* \times [N_\t{C}]$ for $n\in[N_\t{data}]$ be an i.i.d.\ dataset of sequence-class pairs sampled from $P(\v{x}, c)$.
The goal in classification is to learn an estimate of the conditional distribution $P(c|\v{x})$.

Examples include e.g.\ sentiment classification, spam filtering, toxicity classification.

\section{Tokenization: How Text is Represented}\label{sec:token}

In the context of natural language tasks, \emph{tokenization} refers to how a piece of text such as ``My grandma makes the best apple pie.'' is represented as a sequence of vocabulary elements (called \emph{tokens}).

\paragraph{Character-level tokenization.}
One possible choice is to let $V$ be the English alphabet (plus punctuation). 
In the example above, we'd get a sequence of length 36: [`M', `y', `~~', ...].
Character-level tokenization tends to yield very long sequences.

\paragraph{Word-level tokenization.}
Another choice would be to let $V$ consist of all English words (plus punctuation).
In the example above, we'd get a sequence of length 7: [`My ', `grandma ', `makes ', ...].
Word-level tokenization tends to require a very large vocabulary and cannot deal with new words at test time.

\paragraph{Subword tokenization.}
This is the method used in practice nowadays: $V$ is a set of commonly occurring word segments like `cious', `ing', `pre'. 
Common words like `is ' are often a separate token, and single characters are also included in $V$ to ensure all words are expressible. 

There are in fact many ways to do subword tokenization.
One of the simplest and most successful ones is Byte Pair Encoding \cite{Gage:94,Sennrich:16} used in GPT-2 \cite{radford19}.

\paragraph{Final vocabulary and text representation.}
Given a choice of tokenization / vocabulary, each vocabulary element is assigned a unique index in $\{1, 2, \dots, N_\t{V}-3 \}$. 
A number of special tokens are then added to the vocabulary.
The number of special tokens varies, and here we will consider three: 
$\texttt{mask\_token} := N_\t{V}-2$, used in masked language modelling (see \Cref{algo:ETraining}); 
$\texttt{bos\_token} := N_\t{V}-1$, used for representing the beginning of sequence; and
$\texttt{eos\_token} := N_\t{V}$, used for representing the end of sequence.
The complete vocabulary has $N_\t{V} = |V|$ elements.

A piece of text is represented as a sequence of indices (called \emph{token IDs}) corresponding to its (sub)words, preceded by $\texttt{bos\_token}$ and followed by $\texttt{eos\_token}$.

\section{Architectural Components}\label{sec:components} 

The following are the neural network building blocks (functions with learnable parameters) from which transformers are made.
Full architectures featuring these building blocks are presented in the next section.
(By a slight abuse of notation, we identify $V$ with the set $\{1, 2, \dots, N_\t{V}\}$.)

\paragraph{Token embedding.}
The token embedding learns to represent each vocabulary element as a vector in $ℝ^{d_\t{e}}$;
see \Cref{algo:token_embedding}.

\begin{algorithm}[h] 
    \DontPrintSemicolon
    \KwIn{$v∈V\cong[N_\t{V}]$, a token ID.}
    \KwOut{$\v{e}∈ℝ^{d_\t{e}}$, the vector representation of the token.}
    \KwParam{$\m{W_e}∈ℝ^{d_\t{e}×N_\t{V}}$, the token embedding matrix.}
    \Return $\v{e} = \m{W_e}[:,v]$
    \caption{Token embedding.}
    \label{algo:token_embedding}
\end{algorithm}

\paragraph{Positional embedding.}
The positional embedding learns to represent a token's position in a sequence as a vector in $ℝ^{d_\t{e}}$.
For example, the position of the first token in a sequence is represented by a (learned) vector $\m{W_p}[:,1]$, the position of the second token is represented by another (learned) vector $\m{W_p}[:,2]$, etc.
The purpose of the positional embedding is to allow a Transformer to make sense of word ordering; in its absence the representation would be
permutation invariant and the model would perceive sequences as ``bags of words'' instead.

Learned positional embeddings require that input sequence length is at most some fixed number $\ell_{\max}$ (the size of the learned positional embedding matrix must be finite and fixed in advance of training).
An intuitive explanation of how this works can be found at \cite{Alammar:18}.
For pseudocode, see \Cref{algo:pos_embedding}.

Not all transformers make use of \emph{learned} positional embeddings, some use a hard-coded mapping $\m{W_p}:ℕ\toℝ^{d_\t{e}}$ instead \cite{Kernes:21}. Such hard-coded positional embeddings can (theoretically) handle arbitrarily long sequences. 
The original Transformer \cite{vaswani17} uses
\begin{align*}
  \m{W_p}[2i-1,t] ~&=~ \sin(t/\ell_{\max}^{2i/d_\t{e}}), \\
  \m{W_p}[2i,t]   ~&=~ \cos(t/\ell_{\max}^{2i/d_\t{e}}).
\end{align*}
for $0<i≤d_\t{e}/2$.

\begin{algorithm}[h] 
    \DontPrintSemicolon
    \KwIn{$\ell∈[\ell_{\max}]$, position of a token in the sequence.}
    \KwOut{$ \v{e_p}∈ℝ^{d_\t{e}}$, the vector representation of the position.}
    \KwParam{$\m{W_p}∈ℝ^{d_\t{e}×\ell_{\max}}$, the positional embedding matrix.}
    \Return $\v{e_p} = \m{W_p}[:,\ell]$
    \caption{Positional embedding.}
    \label{algo:pos_embedding}
\end{algorithm}

The positional embedding of a token is usually added to the token embedding to form a token's initial embedding.
For the $t$-th token of a sequence $\v{x}$, the embedding is 
\begin{equation}
 \v{e}=\m{W_e}[:,x[t]] + \m{W_p}[:,t].
\end{equation}

\paragraph{Attention.}
Attention is the main architectural component of transformers.
It enables a neural network to make use of contextual information (e.g.\ preceding text or the surrounding text) for predicting the current token.

On a high level, attention works as follows:
the token currently being predicted is mapped to a \emph{query} vector $\v{q}∈ℝ^{d_\t{attn}}$, and the tokens in the context are mapped to \emph{key} vectors $\v{k}_t∈ℝ^{d_\t{attn}}$ and \emph{value} vectors $\v{v}_t∈ℝ^{d_\t{value}}$.
The inner products $\v{q}^\intercal \v{k}_t$ are interpreted as the degree to which token $t∈V$ is important for predicting the current token $q$ -- they are used to derive a distribution over the context tokens, which is then used to combine the value vectors.
An intuitive explanation how this achieves attention can be found at \cite{Alammar:18,Alammar:19}.
The precise algorithm is given in \Cref{algo:basic_attention}.

\begin{algorithm}[h]
    \caption{Basic single-query attention.}
    \label{algo:basic_attention}
    \DontPrintSemicolon
    \KwIn{$\v{e}∈ℝ^{d_\t{in}}$, vector representation of the current token} 
    \KwIn{$\v{e}_t∈ℝ^{d_\t{in}}$, vector representations of context tokens $t∈[T]$.}
    \KwOut{$\v{\tilde v}∈ℝ^{d_\t{out}}$, vector representation of the token and context combined.}
    \KwParam{$\m{W_q}, \m{W_k}∈ℝ^{d_\t{attn}×d_\t{in}}$, $ \v{b_q}, \v{b_k}∈ℝ^{d_\t{attn}}$, the query and key linear projections.}
    \KwParam{$\m{W_v}∈ℝ^{d_\t{out}×d_\t{in}}$, $ \v{b_v}∈ℝ^{d_\t{out}}$, the value linear projection.}
    $\v{q} \gets \m{W_q} \v{e} + \v{b_q}$ \;
    $\forall t:\ \v{k}_t \gets \m{W_k} \v{e}_t + \v{b_k}$ \;
    $\forall t:\ \v{v}_t \gets \m{W_v} \v{e}_t + \v{b_v}$ \;
    $\forall t:\ \alpha_t = \frac{ \exp\left(\v{q}^\intercal \v{k}_t /\sqrt{d_\t{attn}}\right) }{ \sum_u \exp\left(\v{q}^\intercal \v{k}_u /\sqrt{d_\t{attn}}\right)} $ \;
    \Return $\v{\tilde v} = \sum_{t=1}^T \alpha_t \v{v}_t$
\end{algorithm}

There are many ways the basic attention mechanism is used in transformers.
We list some of the most common variants below.

It will be useful to define the softmax function for matrix arguments, 
as well as a Mask matrix:
\begin{equation}
    \t{softmax}( \m{A})[t_\mathrm{z}, t_\mathrm{x}] ~:=~ \frac{\exp A[t_\mathrm{z},t_\mathrm{x}]}{ \sum_{t} \exp A[t,t_\mathrm{x}]},
\end{equation}
\begin{equation}\label{eq:mask}
    \text{Mask}[t_\t{z},t_\t{x}] = \left\{{1~~~ \text{for bidirectional attention} 
                                   \atop [\![t_\t{z}\!≤\!t_\t{x}]\!]~~ \text{for unidirectional att.}}\right.
\end{equation}

\begin{algorithm}[h] 
    \caption{$\m{\tilde V}\gets$~\texttt{Attention}$(\m{X},\m{Z}|\bmcWqkv,\text{Mask})$}
    \label{algo:attention}
    \DontPrintSemicolon
    \tcc{Computes a single (masked) self- or cross- attention head.}
    \KwIn{$\m{X}∈ℝ^{d_\t{x}×\ell_\t{x}}, \m{Z}∈ℝ^{d_\t{z}×\ell_\t{z}}$, vector representations of primary and context  sequence.}
    \KwOut{$\m{\tilde V}∈ℝ^{d_\t{out}×\ell_\t{x}} $, updated representations of tokens in $\m{X}$, folding in information from tokens in $\m{Z}$.}
    \KwParam{$\bmcWqkv$ consisting of: 
            $\m{W_q}∈ℝ^{d_\t{attn}×d_\t{x}}$, $\v{b_q}∈ℝ^{d_\t{attn}}$
             $\m{W_k}∈ℝ^{d_\t{attn}×d_\t{z}}$, $\v{b_k}∈ℝ^{d_\t{attn}}$
             $\m{W_v}∈ℝ^{d_\t{out}×d_\t{z}}$, ~$\v{b_v}∈ℝ^{d_\t{out}}$.}
    \KwHyper{Mask$∈\!\!\{0,\!1\}^{\ell_\t{z}×\ell_\t{x}}$, $\uparrow$\eqref{eq:mask}}
    $\m{Q} \gets \m{W_q} \m{X} + \v{b_q} \v{1}^\intercal$ ~~~ [\![$\m{Q}\t{uery}∈ℝ^{d_\t{attn}×\ell_\t{x}}$]\!] \;
    $\m{K} \gets \m{W_k} \m{Z} + \v{b_k} \v{1}^\intercal$ ~~~~~~~ [\![$\m{K}\t{ey}∈ℝ^{d_\t{attn}×\ell_\t{z}}$]\!] \;
    $\m{V} \gets \m{W_v} \m{Z} + \v{b_v} \v{1}^\intercal$ ~~~~~ [\![$\m{V}\t{alue}∈ℝ^{d_\t{out}×\ell_\t{z}}$]\!] \;
    $\m{S} \gets \m{K}^\intercal \m{Q}$ ~~~~~~~~~~~~~~~~~~~ [\![$\m{S}\t{core}∈ℝ^{\ell_\t{z}×\ell_\t{x}}$]\!] \;
    $\forall t_\t{z}, t_\t{x},$ if $\neg$Mask$[t_\t{z},t_\t{x}]$ then $S[t_\t{z},t_\t{x}] \gets -∞ $ \;
    \Return $\m{\tilde V} = \m{V}\cdot \t{softmax}\left( \m{S} / \sqrt{d_\t{attn}} \right)$    
\end{algorithm}

\paragraph{Bidirectional / unmasked self-attention.}
Given a sequence of token representations, this variant applies attention to each token, treating all tokens in the sequence as the context.
See \Cref{algo:attention}, called with token sequence $\m{Z}=\m{X}$ and no masking (Mask≡1).

\paragraph{Unidirectional / masked self-attention.}
Given a sequence of token representations, this variant applies attention to each token, treating all preceding tokens (including itself) as the context.
Future tokens are masked out, so this causal auto-regressive version can be used for online prediction.
See \Cref{algo:attention}, called with token sequence $\m{Z}=\m{X}$ and Mask$[t_\t{z}, t_\t{x}]:=[\![t_\t{z} \leq t_\t{x}]\!]$.
For this Mask, the output $\m{\tilde V}[:,1:t]$ only depends on $\m{X}[:,1:t]$,
hence can be used to predict $\m{X}[:,t+1]$. 

\paragraph{Cross-attention.}
Given two sequences of token representations (often in the context of a sequence-to-sequence task), 
this variant applies attention to each token of the primary token sequence $\m{X}$, treating the second token sequence $\m{Z}$ as the context.
See \Cref{algo:attention}, called with Mask≡1. 
While the output $\m{\tilde V}$ and input sequences $\m{X}$ have the same length $\ell_\t{x}$, 
the context sequence $\m{Z}$ can have different length $\ell_\t{z}$.

\paragraph{Multi-head attention.}
The attention algorithm presented so far (\Cref{algo:attention}) describes the operation of a single \emph{attention head}.
In practice, transformers run multiple attention heads (with separate learnable parameters) in parallel and combine their outputs; 
this is called \emph{multi-head attention}; see \Cref{algo:MHAttention}

\begin{algorithm}[h] 
    \caption{$\m{\tilde V}\gets$~\texttt{MHAttention}$(\m{X},\m{Z}|\bmcW,\text{Mask})$}
    \label{algo:MHAttention}
    \DontPrintSemicolon
    \tcc{Computes Multi-Head (masked) self- or cross- attention layer.}
    \KwIn{$\m{X}∈ℝ^{d_\t{x}×\ell_\t{x}}, \m{Z}∈ℝ^{d_\t{z}×\ell_\t{z}}$, vector representations of primary and context  sequence.}
    \KwOut{$\m{\tilde V}∈ℝ^{d_\t{out}×\ell_\t{x}} $, updated representations of tokens in $\m{X}$, folding in information from tokens in $\m{Z}$.}
    \KwHyper{$H$, number of attention heads}
    \KwHyper{Mask$∈\!\!\{0,\!1\}^{\ell_\t{z}×\ell_\t{x}}$, $\uparrow$\eqref{eq:mask}}
    \KwParam{$\bmcW$ consisting of \hfill \\
    ~~~For $h∈[H]$, $\bmcWqkv^h$ consisting of: \hfill \\
    ~~~~~$|$~~$\m{W}_\m{q}^h∈ℝ^{d_\t{attn}×d_\t{x}}$, $ \v{b}_\m{q}^h∈ℝ^{d_\t{attn}}$, \\
    ~~~~~$|$~~$\m{W}_\m{k}^h∈ℝ^{d_\t{attn}×d_\t{z}}$, $ \v{b}_\m{k}^h∈ℝ^{d_\t{attn}}$, \\
    ~~~~~$|$~~$\m{W}_\m{v}^h∈ℝ^{d_\t{mid}×d_\t{z}}$, $ \v{b}_\m{v}^h∈ℝ^{d_\t{mid}}$. \\
    ~~~$\m{W_o}∈ℝ^{d_\t{out}× Hd_\t{mid}}$, $ \v{b_o}∈ℝ^{d_\t{out}}$.}
    For $h∈[H]$:\ \\ \hfill $\m{Y}^h \gets \texttt{Attention}(\m{X}, \m{Z} | \bmcWqkv^h, \text{Mask})$ \; 
    $ \m{Y} \gets [\m{Y}^1; \m{Y}^2; \dots ; \m{Y}^H]$ \;
    \Return $\m{\tilde V} = \m{W_o} \m{Y} + \v{b_o}\v{1}^\intercal$
\end{algorithm}

\paragraph{Layer normalisation.}
Layer normalisation explicitly controls the mean and variance of individual neural network activations; 
the pseudocode is given in \Cref{algo:layer_norm}.
Some transformers use a simpler and more computationally efficient version of layer normalization setting $\v{m}=\vbe=0$, called root mean square layer normalisation, or RMSnorm.

\begin{algorithm}[t] 
    \caption{$\hat{\v{e}}\gets$~\texttt{layer\_norm}$(\v{e}|\vga,\vbe)$}
    \label{algo:layer_norm}
    \DontPrintSemicolon
    \tcc{Normalizes layer activations $\v{e}$.}
    \KwIn{$\v{e}∈ℝ^{d_\t{e}}$, neural network activations.}
    \KwOut{$\widehat{\v{e}}∈ℝ^{d_\t{e}} $, normalized activations.}
    \KwParam{$\vga,\vbe∈ℝ^{d_\t{e}}$, element-wise scale and offset.}
    $\v{m} \gets \sum_{i=1}^{d_\t{e}} \v{e}[i] / d_\t{e}$ \;
    $v \gets \sum_{i=1}^{d_\t{e}} (\v{e}[i] - \v{m})^2 /d_\t{e} $ \;
    \Return $ \widehat{\v{e}} = \frac{\v{e} - \v{m}} {\sqrt{v}} ⊙ \vga + \vbe$, where $⊙$ denotes element-wise multiplication.
\end{algorithm}

\paragraph{Unembedding.}
The unembedding learns to convert a vector representation of a token and its context into a distribution over the vocabulary elements; see \Cref{algo:unembedding}.
The algorithm describes an independently learned unembedding matrix, but note that sometimes the unembedding matrix is instead fixed to be the transpose of the embedding matrix.

\begin{algorithm}[t] 
    \DontPrintSemicolon
    \KwIn{$\v{e}∈ℝ^{d_\t{e}}$, a token encoding.}
    \KwOut{$ \v{p}∈\Delta(V)$, a probability distribution over the vocabulary.}
    \KwParam{$\m{W_u}∈ℝ^{N_\t{V}×d_\t{e}}$, the unembedding matrix.}
    \Return $\v{p} = \t{softmax}(\m{W_u}\v{e})$ \;
    \caption{Unembedding.}
    \label{algo:unembedding}
\end{algorithm}

\section{Transformer Architectures}\label{sec:architectures} 

This section presents a few prominent transformer architectures, 
based on attention \Cref{algo:attention,algo:MHAttention} and using normalization \Cref{algo:layer_norm}, in roughly historical order:
\begin{itemize}
    \item EDT \cite{vaswani17} The original sequence-to-sequence / Encoder-Decoder Transformer, 
    \Cref{algo:EDTransformer,algo:EDTraining,algo:EDInference}.
    \item BERT \cite{devlin19}, which is an instance of an encoder-only transformer (encoder-only means that it is derived from the encoder-decoder architecture by dropping the decoder part), 
    \Cref{algo:ETransformer,algo:ETraining}.
    \item GPT \citep{radford19,brown20}, which is an instance of a decoder-only transformer, 
    \Cref{algo:DTransformer,algo:DTraining,algo:DInference}.
\end{itemize}
While the main architectural difference between BERT and GPT is in attention masking, they also differ in a number of less important ways: e.g.\ they use different activation functions and the layer-norms are positioned differently.
We included these differences in the pseudocode to stay faithful to the original algorithms, but note that different transformer architectures may adopt these selectively.

To simplify notation, we denote by $\bmcW$ the entire set of parameters (query, key, value and output linear projections) required by a multi-head attention layer:
\begin{align}\label{eq:Wall}
    \bmcW ~:=~ \left( \begin{matrix}
    \m{W}_\m{q}^h∈ℝ^{d_\t{attn}×d_\t{x}}, & \v{b}_\v{q}^h∈ℝ^{d_\t{attn}}, & 
    h∈[H] \\
    \m{W}_\m{k}^h∈ℝ^{d_\t{attn}×d_\t{z}}, & \v{b}_\v{k}^h∈ℝ^{d_\t{attn}}, & 
    h∈[H] \\
    \m{W}_\m{v}^h∈ℝ^{d_\t{mid}×d_\t{z}},  & \v{b}_\v{v}^h∈ℝ^{d_\t{mid}}, & 
    h∈[H] \\
    \m{W_o}∈ℝ^{d_\t{out}× Hd_\t{mid}},            & \v{b_o}∈ℝ^{d_\t{out}} &
    \end{matrix}\right) 
\end{align}

\paragraph{Encoder-decoder / sequence-to-sequence transformer \citep{vaswani17}.}
This is the very first transformer. 
It was originally used for sequence-to-sequence tasks (machine translation), which is why it is more complicated than its successors.

The idea behind the architecture is as follows:
First, the context sequence is encoded using bidirectional multi-head attention. The output of this `encoder' part of the network is a vector representation of each context token, taking into account the entire context sequence.
Second, the primary sequence is encoded. Each token in the primary sequence is allowed to use information from the encoded context sequence, as well as primary sequence tokens that precede it.
See \Cref{algo:EDTransformer} for more details.

\begin{algorithm*}[h] 
    \caption{$\m{P}\gets$~\texttt{EDTransformer}$(\v{z}, \v{x}|\vth)$}
    \label{algo:EDTransformer}
    \DontPrintSemicolon
    \tcc{Encoder-decoder transformer forward pass}
    \KwIn{$\v{z}, \v{x}∈V^*$, two sequences of token IDs.}
    \KwOut{$ \m{P}∈(0,1)^{N_\t{V}×\t{length}( \v{x})}$, where the $t$-th column of $ \m{P}$ represents $\hat P_{\vth}(x[t+1]|\, \v{x}[1:t], \v{z})$. }
    \KwHyper{$\ell_{\max}, L_\t{enc}, L_\t{dec}, H, d_\t{e}, d_\t{mlp}∈ℕ$}
    \KwParam{$\vth$ includes all of the following parameters: \;
    ~~~$\m{W_e}∈ℝ^{d_\t{e}×N_\t{V}}$, $\m{W_p}∈ℝ^{d_\t{e}×\ell_{\max}}$, the token and positional embedding matrices. \;
    ~~~For $l∈[L_\t{enc}]$: \;
    ~~~~~$|$~~$\bmcW_l^\t{enc}$, multi-head encoder attention parameters for layer $l$, see \eqref{eq:Wall}, \;
    ~~~~~$|$~~$\vga_l^1, \vbe_l^1, \vga_l^2, \vbe_l^2∈ℝ^{d_\t{e}}$,\ \ two sets of layer-norm parameters, \;
    ~~~~~$|$~~$\m{W}_\t{mlp1}^l∈ℝ^{d_\t{mlp}×{d_\t{e}}}$, 
                $\v{b}_\t{mlp1}^l∈ℝ^{d_\t{mlp}}$,
                $\m{W}_\t{mlp2}^l∈ℝ^{d_\t{e}×d_\t{mlp}}$, 
                $\v{b}_\t{mlp2}^l∈ℝ^{d_\t{e}}$,\ \ MLP parameters. \;
    ~~~For $l∈[L_\t{dec}]$: \;
    ~~~~~$|$~~$\bmcW_l^\t{dec}$, multi-head decoder attention parameters for layer $l$, see \eqref{eq:Wall}, \;
    ~~~~~$|$~~$\bmcW_l^\t{e/d}$, multi-head cross-attention parameters for layer $l$, see \eqref{eq:Wall}, \;
    ~~~~~$|$~~$\vga_l^3, \vbe_l^3, \vga_l^4, \vbe_l^4, \vga_l^5, \vbe_l^5∈ℝ^{d_\t{e}}$,\ \ three sets of layer-norm parameters, \;
    ~~~~~$|$~~$\m{W}_\t{mlp3}^l∈ℝ^{d_\t{mlp}×d_\t{e}}$, 
                $\v{b}_\t{mlp3}^l∈ℝ^{d_\t{mlp}}$,
                $\m{W}_\t{mlp4}^l∈ℝ^{d_\t{e}×d_\t{mlp}}$, 
                $\v{b}_\t{mlp4}^l∈ℝ^{d_\t{e}}$,\ \ MLP parameters.  \;
    ~~~$ \m{W_u}∈ℝ^{N_\t{V}×d_\t{e}}$, the unembedding matrix.}
    \tcc{Encode the context sequence:}
    $\ell_\t{z} \gets \t{length}(\v{z})$ \;
    for $t∈[\ell_\t{z}]:\ \v{e}_t \gets \m{W_e}[:,z[t]] + \m{W_p}[:,t]$ \;
    $ \m{Z} \gets [\v{e}_1, \v{e}_2, \dots \v{e}_{\ell_\t{z}}]$ \;
    \For{$l = 1, 2, \dots, L_\t{enc}$}{
        $ \m{Z} \gets \m{Z} + \texttt{MHAttention}(\m{Z}\,|\bmcW_l^\t{enc}, \text{Mask}≡1)$ \; 
        for $t∈[\ell_\t{z}]:\ \m{Z}[:,t] \gets \texttt{layer\_norm}(\m{Z}{[:,t]}\,|\,\vga_l^1, \vbe_l^1)$ \;
        $ \m{Z} \gets \m{Z} + \m{W}_\t{mlp2}^l \texttt{ReLU}(\m{W}_\t{mlp1}^l \m{Z} + \v{b}_\t{mlp1}^l \v{1}^\intercal) + \v{b}_\t{mlp2}^l \v{1}^\intercal$ \;
        for $t∈[\ell_\t{z}]:\ \m{Z}[:,t] \gets \texttt{layer\_norm}(\m{Z}{[:,t]}\,|\,\vga_l^2, \vbe_l^2)$ \;
    }
    \tcc{Decode the primary sequence, conditioning on the context:}
    $\ell_\t{x} \gets \t{length}(\v{x})$ \;
    for $t∈[\ell_\t{x}]:\ \v{e}_t \gets \m{W_e}[:,x[t]] + \m{W_p}[:,t]$ \;
    $ \m{X} \gets [\v{e}_1, \v{e}_2, \dots \v{e}_{\ell_\t{x}}]$ \;
    \For{$i = 1, 2, \dots, L_\t{dec}$}{
        $ \m{X} \gets \m{X} + \texttt{MHAttention}(\m{X}\,|\bmcW_l^\t{dec}, \text{Mask}[t, t']≡ [\![t \leq t' ]\!] )$ \; 
        for $t∈[\ell_\t{x}]:\ \m{X}[:,t] \gets \texttt{layer\_norm}(\m{X}{[:,t]}\,|\,\vga_l^3, \vbe_l^3)$ \;
        $ \m{X} \gets \m{X} + \texttt{MHAttention}(\m{X}, \m{Z}\,|\bmcW_l^\t{e/d}, \text{Mask}≡1)$ \; 
        for $t∈[\ell_\t{x}]:\ \m{X}[:,t] \gets \texttt{layer\_norm}(\m{X}{[:,t]}\,|\,\vga_l^4, \vbe_l^4)$ \;
        $ \m{X} \gets \m{X} + \m{W}_\t{mlp4}^l \texttt{ReLU}(\m{W}_\t{mlp3}^l \m{X} + \v{b}_\t{mlp3}^l \v{1}^\intercal) + \v{b}_\t{mlp4}^l \v{1}^\intercal$ \;
        for $t∈[\ell_\t{x}]:\ \m{X}[:,t] \gets \texttt{layer\_norm}(\m{X}{[:,t]}\,|\,\vga_l^5, \vbe_l^5)$ \;
    }
    \tcc{Derive conditional probabilities and return:}
    \Return $\m{P} = \t{softmax}( \m{W_u}\m{X}) $
\end{algorithm*}

\paragraph{Encoder-only transformer: BERT \citep{devlin19}.}
BERT is a bidirectional transformer trained on the task of masked language modelling.
Given a piece of text with some tokens masked out, the goal is to correctly recover the masked-out tokens.
The original use of BERT was to learn generally useful text representations, which could then be adapted for various downstream NLP tasks.
The masking is not performed via the Mask parameter but differently:
During training each input token is replaced with probability $p_\text{mask}$ by a dummy token \texttt{mask\_token},
and evaluation is based on the reconstruction probability of these knocked-out tokens (see \Cref{algo:ETraining}).

The BERT architecture resembles the encoder part of the seq2seq transformer (hence `encoder-only'). It is described in detail in \Cref{algo:ETransformer}.
It uses the GELU nonlinearity instead of ReLU:
\begin{equation}
    \t{GELU}(x) = x \cdot ℙ_{X\sim \mathcal{N}(0,1)}[X < x].
\end{equation}
(When called with vector or matrix arguments, GELU is applied element-wise.)

\begin{algorithm*}[h] 
    \caption{$\m{P}\gets$~\texttt{ETransformer}($\v{x}|\vth)$}
    \label{algo:ETransformer}
    \DontPrintSemicolon
    \tcc{BERT, an encoder-only transformer, forward pass}
    \DontPrintSemicolon
    \KwIn{$\v{x}∈V^*$, a sequence of token IDs.}
    \KwOut{$ \m{P}∈(0,1)^{N_\t{V}×\t{length}( \v{x})}$, where each column of $ \m{P}$ is a distribution over the vocabulary. }
    \KwHyper{$\ell_{\max}, L, H, d_\t{e}, d_\t{mlp}, d_\t{f}∈ℕ$}
    \KwParam{$\vth$ includes all of the following parameters: \;
    ~~~$\m{W_e}∈ℝ^{d_\t{e}×N_\t{V}}$, $\m{W_p}∈ℝ^{d_\t{e}×\ell_{\max}}$, the token and positional embedding matrices. \;
    ~~~For $l∈[L]$: \;
    ~~~~~$|$~~$\bmcW_l$, multi-head attention parameters for layer $l$, see \eqref{eq:Wall}, \;
    ~~~~~$|$~~$\vga_l^1, \vbe_l^1, \vga_l^2, \vbe_l^2∈ℝ^{d_\t{e}}$,\ \ two sets of layer-norm parameters, \;
    ~~~~~$|$~~$\m{W}_\t{mlp1}^l∈ℝ^{d_\t{mlp}×d_\t{e}}$, 
                $\v{b}_\t{mlp1}^l∈ℝ^{d_\t{mlp}}$,
                $\m{W}_\t{mlp2}^l∈ℝ^{d_\t{e}×d_\t{mlp}}$, 
                $\v{b}_\t{mlp2}^l∈ℝ^{d_\t{e}}$,\ \ MLP parameters. \;
    ~~~$\m{W_f}∈ℝ^{d_\t{f}×d_\t{e}}, \v{b_f}∈ℝ^{d_\t{f}}$, $\vga,\vbe∈ℝ^{d_\t{f}}$, the final linear projection and layer-norm parameters. \;
    ~~~$ \m{W_u}∈ℝ^{N_\t{V}×d_\t{e}}$, the unembedding matrix.}
    $\ell \gets \t{length}(\v{x})$ \;
    for $t∈[\ell]:\ \v{e}_t \gets \m{W_e}[:,x[t]] + \m{W_p}[:,t]$ \;
    $ \m{X} \gets [\v{e}_1, \v{e}_2, \dots \v{e}_{\ell}]$ \;
    \For{$l = 1, 2, \dots, L$}{
        $ \m{X} \gets \m{X} + \texttt{MHAttention}(\m{X}\,|\bmcW_l, \text{Mask}≡1)$ \; 
        for $t∈[\ell]:\ \m{X}[:,t] \gets \texttt{layer\_norm}(\m{X}{[:,t]}\,|\,\vga_l^1, \vbe_l^1)$ \;
        $ \m{X} \gets \m{X} + \m{W}_\t{mlp2}^l \texttt{GELU}(\m{W}_\t{mlp1}^l \m{X} + \v{b}_\t{mlp1}^l \v{1}^\intercal) + \v{b}_\t{mlp2}^l \v{1}^\intercal$ \;
        for $t∈[\ell]:\ \m{X}[:,t] \gets \texttt{layer\_norm}(\m{X}{[:,t]}\,|\,\vga_l^2, \vbe_l^2)$ \;
    }
    $\m{X} \gets \texttt{GELU}(\m{W_f} \m{X} + \v{b_f}\v{1}^\intercal)$  \;
    for $t∈[\ell]:\ \m{X}[:,t] \gets \texttt{layer\_norm}(\m{X}{[:,t]}\,|\,\vga, \vbe)$ \;
    \Return $\m{P} = \t{softmax}( \m{W_u}\m{X}) $
\end{algorithm*}

\paragraph{Decoder-only transformers: GPT-2 \citep{radford19}, GPT-3 \citep{brown20}, Gopher \citep{rae21}.}
GPT-2 and GPT-3 are large language models developed by OpenAI, and Gopher is a large language model developed by DeepMind.
They all have similar architectures and are trained by autoregressive language modelling: Given an incomplete sentence or paragraph, the goal is to predict the next token.

The main difference from BERT is that GPT/Gopher use unidirectional attention instead of bidirectional attention; 
they also apply layer-norms in slightly different order.

See \Cref{algo:DTransformer} for the pseudocode of \mbox{GPT-2}.
GPT-3 is identical except larger, and replaces dense attention in Line 6 by sparse attention, i.e.\ each token only uses a subset of the full context.

Gopher also deviates only slightly from the \mbox{GPT-2} architecture: it replaces layer norm in lines 5, 7 and 10 by RMSnorm ($m=\vbe=0$), and it uses different positional embeddings.

\begin{algorithm*}[h] 
    \caption{$\m{P}\gets$~\texttt{DTransformer}($\v{x}|\vth)$}
    \label{algo:DTransformer}
    \DontPrintSemicolon
    \tcc{GPT, a decoder-only transformer, forward pass}
    \KwIn{$\v{x}∈V^*$, a sequence of token IDs.}
    \KwOut{$ \m{P}∈(0,1)^{N_\t{V}×\t{length}( \v{x})}$, where the $t$-th column of $ \m{P}$ represents $\hat P_{\vth}(x[t+1]|\, \v{x}[1:t])$. }
    \KwHyper{$\ell_{\max}, L, H, d_\t{e}, d_\t{mlp}∈ℕ$}
    \KwParam{$\vth$ includes all of the following parameters: \;
    ~~~$\m{W_e}∈ℝ^{d_\t{e}×N_\t{V}}$, $\m{W_p}∈ℝ^{d_\t{e}×\ell_{\max}}$, the token and positional embedding matrices. \;
    ~~~For $l∈[L]$: \;
    ~~~~~$|$~~$\bmcW_l$, multi-head attention parameters for layer $l$, see \eqref{eq:Wall}, \;
    ~~~~~$|$~~$\vga_l^1, \vbe_l^1, \vga_l^2, \vbe_l^2∈ℝ^{d_\t{e}}$,\ \ two sets of layer-norm parameters, \;
    ~~~~~$|$~~$\m{W}_\t{mlp1}^l∈ℝ^{d_\t{mlp}×d_\t{e}}$, 
                $\v{b}_\t{mlp1}^l∈ℝ^{d_\t{mlp}}$,
                $\m{W}_\t{mlp2}^l∈ℝ^{d_\t{e}×d_\t{mlp}}$, 
                $\v{b}_\t{mlp2}^l∈ℝ^{d_\t{e}}$,\ \ MLP parameters. \;
    ~~~$\vga,\vbe∈ℝ^{d_\t{e}}$, final layer-norm parameters. \;
    ~~~$ \m{W_u}∈ℝ^{N_\t{V}×d_\t{e}}$, the unembedding matrix.}
    $\ell \gets \t{length}(\v{x})$ \;
    for $t∈[\ell]:\ \v{e}_t \gets \m{W_e}[:,x[t]] + \m{W_p}[:,t]$ \;
    $ \m{X} \gets [\v{e}_1, \v{e}_2, \dots \v{e}_{\ell}]$ \;
    \For{$l = 1, 2, \dots, L$}{
        for $t∈[\ell]:\ \m{\tilde X}[:,t] \gets \texttt{layer\_norm}(\m{X}{[:,t]}\,|\,\vga_l^1, \vbe_l^1)$ \;
        $ \m{X} \gets \m{X} + \texttt{MHAttention}(\m{\tilde X}\,|\bmcW_l, \text{Mask}[t, t']=[\![t \leq t']\!])$ \; 
        for $t∈[\ell]:\ \m{\tilde X}[:,t] \gets \texttt{layer\_norm}(\m{X}{[:,t]}\,|\,\vga_l^2, \vbe_l^2)$ \;
        $ \m{X} \gets \m{X} + \m{W}_\t{mlp2}^l \texttt{GELU}(\m{W}_\t{mlp1}^l \m{\tilde X} + \v{b}_\t{mlp1}^l \v{1}^\intercal) + \v{b}_\t{mlp2}^l \v{1}^\intercal$ \;
    }
    for $t∈[\ell]:\ \m{X}[:,t] \gets \texttt{layer\_norm}(\m{X}{[:,t]}\,|\,\vga, \vbe)$ \;
    \Return $\m{P} = \t{softmax}( \m{W_u}\m{X}) $
\end{algorithm*}

\paragraph{Multi-domain decoder-only transformer: Gato \citep{reed22}.}
Gato is a multi-modal multi-task transformer built by DeepMind.
It is a single neural network that can play Atari, navigate 3D environments, control a robotic arm, caption images, have conversations, and more.

Under the hood, each modality is converted into a sequence prediction problem by a separate tokenization and embedding method;
for example images are divided into non-overlapping $16×16$ patches, ordered in raster order (left-to-right, top-to-bottom) and processed by a ResNet block to obtain a vector representation.

The actual Gato architecture is then a decoder-only transformer like the one in \Cref{algo:DTransformer}, but where Line 2 is replaced with modality-specific embedding code.

\section{Transformer Training and Inference}\label{sec:train}\label{sec:inf}

This section lists the pseudocode for various algorithms for training and using transformers:
\begin{itemize}
    \item {\bf EDTraining()} \Cref{algo:EDTraining} shows how to train a sequence-to-sequence transformer (the original Transformer \citep{vaswani17}).
    \item {\bf ETraining()} \Cref{algo:ETraining} shows how to train a transformer on the task of masked language modelling (like BERT \citep{devlin19}).
    \item {\bf DTraining()} \Cref{algo:DTraining} shows how to train a transformer on the task of next token prediction (like CPT-x \citep{brown20} and Gopher \citep{rae21}).
    \item {\bf DInference()} \Cref{algo:DInference} shows how to prompt a transformer trained on next token prediction (like GPT-x \citep{brown20}).
    The temperature parameter $τ$ interpolates between most likely continuation ($τ=0$), faithful sampling ($τ=1)$, and uniform sampling $(τ=∞)$.
    \item {\bf EDInference()} \Cref{algo:EDInference} shows how to use a sequence-to-sequence transformer for prediction.
\end{itemize}

\paragraph{Gradient descent.}
The described training \Crefrange{algo:EDTraining}{algo:DTraining}
use Stochastic Gradient Descent (SGD) $\vth\gets \vth- η\cdot  ∇\text{loss}(\vth)$ to minimize the log loss (aka cross entropy) as the update rule.
Computation of the gradient is done via automatic differentiation tools; see \citep[Table~5]{Baydin:18}.
In practice, vanilla SGD is usually replaced by some more refined variation such as RMSProp or AdaGrad or others \cite{Ruder:16}. 
Adam \cite{kingma14} is used most often these days.

\begin{algorithm}[h] 
    \caption{$\hat{\vth}\gets$~\texttt{EDTraining}($\v{z}_{1:N_\t{data}}, \v{x}_{1:N_\t{data}},\vth$)}
    \label{algo:EDTraining}
    \DontPrintSemicolon
    \tcc{Training a seq2seq model}
    \KwIn{$\{ (\v{z}_n, \v{x}_n) \}_{n=1}^{N_\t{data}}$, a dataset of sequence pairs.}
    \KwIn{$\vth$, initial transformer parameters.}
    \KwOut{$\hat{\vth}$, the trained parameters.}
    \KwHyper{$N_\t{epochs}∈ℕ,\ η∈(0,∞)$}
    \For{$i = 1, 2, \dots, N_\t{epochs}$}{
        \For{$n = 1, 2, \dots N_\t{data}$}{
            $\ell \gets \t{length}(\v{x}_n)$ \;
            $ \m{P}(\vth) \gets \texttt{EDTransformer}(\v{z}_n, \v{x}_n | \vth)\!\!$ \;
            $\t{loss}(\vth) = -\sum_{t=1}^{\ell-1} \log P(\vth)[x_n[t\!+\!1],t]\!\!$ \;
            $\vth \gets \vth - η \cdot \nabla \t{loss}(\vth) $ \;
        }
    }
    \Return $\hat{\vth} = \vth$
\end{algorithm}

\begin{algorithm}[h] 
    \caption{$\hat{\vth}\gets$~\texttt{ETraining}($\v{x}_{1:N_\t{data}},\vth$)}
    \label{algo:ETraining}
    \DontPrintSemicolon
    \tcc{Training by masked language modelling}
    \KwIn{$\{ \v{x}_n \}_{n=1}^{N_\t{data}}$, a dataset of sequences.}
    \KwIn{$\vth$, initial encoder-only transformer parameters.}
    \KwOut{$\hat{\vth}$, the trained parameters.}
    \KwHyper{$N_\t{epochs}∈ℕ,\ η∈(0,∞), p_\t{mask}∈(0, 1)$}
    \For{$i = 1, 2, \dots, N_\t{epochs}$}{
        \For{$n = 1, 2, \dots, N_\t{data}$}{
            $\ell \gets \t{length}(\v{x}_n)$ \;
            \For{$t = 1, 2, \dots, \ell$}{
                $\tilde x_n[t] \gets \texttt{mask\_token} \text{ or } x_n[t]$ randomly with probability $p_\t{mask}$ or $1-p_\t{mask}$ \;
            }
            $\tilde T \!\gets\! {\{t\!∈\![\ell]: \tilde x_n[t] = \texttt{mask\_token}\}}\!\!$ \;
            $ \m{P}(\vth) \gets \texttt{ETransformer}(\v{\tilde x}_n \,|\, \vth)$ \;
            $\t{loss}(\vth) = -\sum_{t∈\tilde T} \log P(\vth)[x_n[t],t] $ \;
            $\vth \gets \vth - η \cdot \nabla \t{loss}(\vth) $ \;
        }
    }
    \Return $\hat{\vth} = \vth$
\end{algorithm}

\begin{algorithm}[h] 
    \caption{$\hat{\vth}\gets$~\texttt{DTraining}($\v{x}_{1:N_\t{data}},\vth$)}
    \label{algo:DTraining}
    \tcc{Training next token prediction}
    \DontPrintSemicolon
    \KwIn{$\{ \v{x}_n \}_{n=1}^{N_\t{data}}$, a dataset of sequences.}
    \KwIn{$\vth$, initial decoder-only transformer parameters.}
    \KwOut{$\hat{\vth}$, the trained parameters.}
    \KwHyper{$N_\t{epochs}∈ℕ,\ η∈(0,∞)$}
    \For{$i = 1, 2, \dots, N_\t{epochs}$}{
        \For{$n = 1, 2, \dots N_\t{data}$}{
            $\ell \gets \t{length}(\v{x}_n)$ \;
            $ \m{P}(\vth) \gets \texttt{DTransformer}(\v{x}_n \,|\, \vth)$ \;
            $\t{loss}(\vth) = -\sum_{t=1}^{\ell-1} \log P(\vth)[x_n[t\!+\!1],t]\!\!$ \;
            $\vth \gets \vth - η \cdot \nabla \t{loss}(\vth) $ \;
        }
    }
    \Return $\hat{\vth} = \vth$
\end{algorithm}

\begin{algorithm}[h] 
    \caption{$\v{y}\gets$~\texttt{DInference}($\v{x},\hat{\vth}$)}
    \label{algo:DInference}
    \DontPrintSemicolon
    \tcc{Prompting a trained model and using it for prediction.}
    \KwIn{Trained transformer parameters $\hat{\vth}$.}
    \KwIn{$\v{x}∈V^*$, a prompt.}    
    \KwOut{$\v{y}∈V^*$, the transformer's continuation of the prompt.}
    \KwHyper{$\ell_\t{gen}∈ℕ,\ τ∈(0,∞)$}
    $\ell \gets \t{length}( \v{x})$ \;
    \For{$i = 1, 2, \dots \ell_\t{gen}$}{
        $ \m{P} \gets \texttt{DTransformer}(\v{x} \,|\, \hat{\vth})$ \;
        $\v{p} \gets \m{P}[:,\ell+i-1]$ \;
        sample a token $y$ from $\v{q} \propto \v{p}^{1/τ}$ \;
        $ \v{x} \gets [\v{x} , y]$ \;
    }
    \Return $\v{y} = \v{x}[\ell+1:\ell+\ell_\t{gen}]$
\end{algorithm}

\begin{algorithm}[h] 
    \caption{$\v{\hat x}\gets$~\texttt{EDInference}($\v{z},\hat{\vth}$)}
    \label{algo:EDInference}
    \DontPrintSemicolon
    \tcc{Using a trained seq2seq model for prediction.}
    \KwIn{A seq2seq transformer and trained parameters $\hat{\vth}$ of the transformer.}
    \KwIn{$ \v{z}∈V^*$, input sequence, e.g.\ a sentence in English.}    
    \KwOut{$\v{\hat x}∈V^*$, output sequence, e.g.\ the sentence in German.}
    \KwHyper{$τ∈(0,∞)$}
    $\v{\hat x} \gets [\texttt{bos\_token}]$ \;
    $y \gets 0$ \;
    \While{$y \neq \texttt{eos\_token}$}{
        $ \m{P} \gets \texttt{EDTransformer}(\v{z}, \v{\hat x} \,|\, \hat{\vth})$ \;
        $\v{p} \gets \m{P}[:, \t{length}(\v{\hat x)}]$ \;
        sample a token $y$ from $\v{q} \propto \v{p}^{1/τ}$ \;
        $ \v{\hat x} \gets [\v{\hat x} , y]$ \;
    }
    \Return $\v{\hat x}$
\end{algorithm}

\section{Practical Considerations}\label{sec:practice}

While the vanilla transformers provided here may work in practice, a variety of ``tricks'' have been developed over the years to improve the performance of deep neural networks in general and transformers in particular \citep{Lin:21}:
\begin{itemize}
\item {\bf Data preprocessing:} cleaning, augmentation \cite{Feng:21}, adding noise, shuffling \citep{Lemke:21} (besides tokenization and chunking). 
\item {\bf Architecture:} sparse layers, weight sharing (besides attention).
\item {\bf Training:} improved optimizers, minibatches, batch  normalization, learning rate scheduling, weight initialization, pre-training, ensembling, multi-task, adversarial (besides layer normalization) \citep{Sutskever:15}. 
\item {\bf Regularization:} weight decay, early stopping, cross-validation, dropout, adding noise \cite{Moradi:20,Tian:22}.
\item {\bf Inference:} scratchpad prompting, few-shot prompting, chain of thought, majority voting \citep{Lewkowycz:22}.
\item {\bf Others.} 
\end{itemize}

\appendix
\section{References}\label{sec:refs}
\def\refname{\vspace{-4ex}}
\bibliographystyle{alpha}
\newcommand{\etalchar}[1]{$^{#1}$}

\begin{table*}
\section{List of Notation}\label{sec:notation}

\begin{tabbing}
  \hspace{0.13\textwidth} \= \hspace{0.13\textwidth}\= \hspace{0.60\textwidth} \= \kill
  {\bf Symbol }      \> {\bf Type}     \> {\bf Explanation}                                      \\[0ex]
  $[N]$              \> $:=\{1,...,N\}$  \> set of integers $1,2,...,N-1,N$                  \\[0ex]
  $i,j$              \> $∈ℕ$           \> generic integer indices       \\[0ex]  
  $V$                \> $\cong[N_\t{V}]$      \> vocabulary                               \\[0ex]
  $N_\t{V}$          \> $∈\mathbb{N}$  \> vocabulary size                                 \\[0ex]
  $V^*$              \> $=\bigcup_{\ell=0}^∞ V^{\ell}$ \> set of token sequences; elements include e.g.\ sentences or documents  \\[0ex]
  $\ell_{\max}$      \> $∈\mathbb{N}$     \> maximum sequence length                          \\[0ex]
  $\ell$             \> $∈[\ell_{\max}]$      \> length of token sequence                           \\[0ex]  
  $t$                \> $∈[\ell]$           \> index of token in a sequence                       \\[0ex]
  $d_{...}$          \> $∈ℕ$           \> dimension of various vectors                       \\[0ex]  
  $\v{x}$            \> $≡x[1:\ell]$   \> $≡x[1]x[2]...x[\ell]∈V^\ell$ ~~ primary token sequence  \\[0ex]
  $\v{z}$            \> $≡z[1:\ell]$   \> $≡z[1]z[2]...z[\ell]∈V^\ell$ ~~ context token sequence  \\[0ex]
  $M[i,j]$           \> $∈ℝ$            \> entry $M_{ij}$ of matrix $M∈ℝ^{d×d'}$        \\[0ex]
  $M[i,:]\!≡\!M[i]$  \> $∈ℝ^{d'}$      \> $i$-th row of matrix $M∈ℝ^{d×d'}$          \\[0ex]
  $M[:,j]$           \> $∈ℝ^d$         \> $j$-th column of matrix $M∈ℝ^{d×d'}$        \\[0ex]
  $\v{e}$            \> $∈ℝ^{d_\t{e}}$    \> vector representation / embedding of a token            \\[0ex]
  $\m{X}$              \> $∈ℝ^{d_\t{e} × \ell_x}$    \> encoded primary token sequence                             \\[0ex]
  $\m{Z}$              \> $∈ℝ^{d_\t{e} × \ell_z}$     \> encoded context token sequence                             \\[0ex]  
  $\t{Mask}$         \> $∈ℝ^{\ell_z\times \ell_x}$   \> masking matrix, it determines the attention context for each token \\[0ex]
  $L, L_\t{enc}, L_\t{dec}$ \> $∈ℕ$       \> number of network (encoder, decoder) layers                   \\[0ex]  
  $l$                \> $∈[L]$           \> index of network layer                             \\[0ex]
  $H$                \> $∈\mathbb{N}$     \> number of attention heads                        \\[0ex]
  $h$                \> $∈[H]$         \> index of attention head                            \\[0ex]
  $N_\t{data}$       \> $∈ℕ$           \> (i.i.d.) sample size                               \\[0ex]  
  $n$                \> $∈[N_\t{data}]$     \> index of sample sequence                         \\[0ex]
  $η$                \> $∈(0,∞)$           \> learning rate                                      \\[0ex]
  $τ$                \> $∈(0,∞)$           \> temperature; it controls the diversity-plausibility trade-off at inference  \\[0ex]
  $\m{W_e}$     \> $∈ℝ^{d_\t{e}×N_\t{V}}$ \> token embedding matrix       \\[0ex]  
  $\m{W_p}$     \> $∈ℝ^{d_\t{e}×\ell_{\max}}$ \> positional embedding matrix       \\[0ex]  
  $\m{W_u}$     \> $∈ℝ^{N_\t{V}×d_\t{e}}$ \> unembedding matrix       \\[0ex]  
  $\m{W_q}$     \> $∈ℝ^{d_\t{attn}×d_\t{x}}$ \> query weight matrix \\[0ex]
  $\v{b_q}$     \> $∈ℝ^{d_\t{attn}}$ \> query bias \\[0ex]
  $\m{W_k}$     \> $∈ℝ^{d_\t{attn}×d_\t{z}}$ \> key weight matrix \\[0ex]
  $\v{b_k}$     \> $∈ℝ^{d_\t{attn}}$ \> key bias \\[0ex]
  $\m{W_v}$     \> $∈ℝ^{d_\t{out}×d_\t{z}}$ \> value weight matrix \\[0ex]
  $\v{b_v}$     \> $∈ℝ^{d_\t{out}}$ \> value bias \\[0ex]
  $\bmcWqkv$     \>  \> collection of above parameters of a single-head attention layer \\[0ex]  
  $\m{W_o}$     \> $∈ℝ^{d_\t{out}×Hd_\t{mid}}$ \> output weight matrix \\[0ex]
  $\v{b_o}$     \> $∈ℝ^{d_\t{out}}$ \> output bias \\[0ex]
  $\bmcW$       \>  \> collection of above parameters of a multi-head attention layer, see \cref{eq:Wall} \\[0ex]
  $\m{W_\t{mlp}}$    \> $∈ℝ^{d_1×d_2}$ \> weight matrix corresponding to an MLP layer in a Transformer \\[0ex]
  $\v{b_\t{mlp}}$    \> $∈ℝ^{d_1}$ \> bias corresponding to an MLP layer in a Transformer \\[0ex]  
  $\vga$            \> $∈ℝ^{d_\t{e}}$ \> layer-norm learnable scale parameter \\[0ex]
  $\vbe$                \> $∈ℝ^{d_\t{e}}$  \> layer-norm learnable offset parameter \\[0ex]
  $\vth,\hat{\vth}$        \> $∈ℝ^d$        \> collection of all learnable / learned Transformer parameters \\[0ex]
\end{tabbing}
\end{table*}

\end{document}